# Maximum Likelihood Joint Tracking and Association in a Strong Clutter

Leonid I. Perlovsky, *Senior Member, IEEE*, and Ross W. Deming, *Member, IEEE*

*Abstract*—We have developed an efficient algorithm for the maximum likelihood joint tracking and association problem in a strong clutter for GMTI data. By using an iterative procedure of the dynamic logic process "from vague-to-crisp," the new tracker overcomes combinatorial complexity of tracking in highly-cluttered scenarios and results in a significant improvement in signal-to-clutter ratio.

## I. INTRODUCTION

Performance of the state-of-the-art algorithms for tracking and association in strong clutter [1], as a function of Signal-to-Clutter Ratio (SCR), is significantly below the information-theoretic limit as indicated by the Cramer-Rao Bound (CRB) for tracking in clutter [2]. The reason for this underperformance is combinatorial complexity of algorithms. When clutter is strong, so that signals are below clutter, multiple associations between data and tracks have to be considered. The number of associations grows combinatorially with the number of data points. Therefore performance is limited by complexity of computations rather than by information in the data. Here we describe a non-combinatorial solution of the maximum likelihood (ML) joint tracking and association problem resulting in a significantly improved performance; it follows a discussion at [3, see 4]. The novel contribution of this paper includes the ML formulation of joint detection, association, and tracking problem, its non-combinatorial solution, and comparison to published performance of other tracker results.

Standard algorithms (such as Multiple Hypotheses Testing, MHT [5]) used in the current GMTI detection and tracking subsystems operate in a two-step process. First, Doppler peaks are detected that exceed a predetermined threshold. Second, these potential target peaks are used to initiate tracks. This two-step procedure is a state-of-the-art approach which is currently used by most tracking systems. The limitation of this procedure is determined by the detection threshold. If the threshold is reduced, the number of detected peaks grows quickly. Increased computer power does not help because the processing requirements are combinatorial in terms of the number of peaks, so that a tenfold increase in the number of peaks results in a billion fold increase in the required computer power. Therefore developing algorithms capable of non-combinatorial detection, association, and tracking is a desirable goal. Before describing such an algorithm below, we briefly review previous development.

A fundamental idea of probabilistic association has originated from Bar-Shalom JPDAF, [6,7], which is more efficient than MHT because one only needs to evaluate the association probabilities separately at each time step. However, since all data association mappings are not considered, this method is not optimal [8,9]. Also, JPDAF is generally not used for track initialization since detection is performed separately from tracking, whereas optimum performance is only obtained when detection and tracking are performed concurrently [2,10]. Both MHT and JPDAF have been adapted for multi-sensor scenarios [7].

In the approach described here, data association is performed concurrently with detection and tracking, without a combinatorial explosion in computational requirements. Also, the approach is appropriate for handling low S/C data and data with multiple, closely-spaced and overlapping targets. Whereas MHT and JPDAF are based upon "hard" associations between measurements and targets, i.e., selecting the best data-to-target mapping to calculate the track parameters, the method described here associates data with target and clutter components probabilistically. Here, the track parameters for a particular target are computed using a weighted combination of all data samples, where the weights correspond to the association probabilities. An advantage of this framework is that it allows an efficient "hill climbing" optimization, to be performed in the space of all model parameters and all possible mappings between data samples and targets. In fact, this procedure can be framed as a special case of the expectation-maximization (EM) algorithm [11,12,13], which has been derived in [14]. Like MHT, this algorithm is optimal [2,14]. However, whereas the computational complexity of MHT scales exponentially with the number of targets, data samples, and sensors, the method described here scales only linearly.

The general framework for the approach described in this paper was developed by Perlovsky in his work on multi-target tracking in clutter, automatic target recognition (ATR), and sensor fusion [2,10,15,16,17,18,19,20,21,22,23], (and other references contained in [14]). A neural network implementation of these ideas for tracking can be found in [[24]].

In these references it has been recognized that if classification features are available, an optimum algorithm will perform target classification concurrently with detection and tracking, by mathematically placing the tracking coordinates (range, Doppler, bearing) on an equal footing with classification features. Thus, the composite model of the data is expressed as a mixture of target and clutter components in the combined space of tracking and classification features. The algorithm described here is based upon this general approach [14] which we refer to as dynamic logic (DL).

The DL tracker has much in common with a similar algorithm known as the probabilistic multihypothesis tracker (PMHT) [8,9,25,26], which was developed independently by Streit and Luginbuhl, and shares strong similarities with Avitzour's approach [27]. The major similarity between DL and these other trackers is the use of EM to perform data association probabilistically, while simultaneously estimating track parameters, and yet, the manner in which EM is used is different, DL does not treat associations as missing variables. It should also be noted that both PMHT [28,29,30] and DL [14,31] have been adapted for combining data from multiple sensors. A difference between DL and PMHT is in the choice of track model. In PMHT the motion of each target is modeled using a set of discrete-time state transition equations, and therefore a Kalman smoother is used to estimate track parameters in the M-step of the EM iterations [8,[25]]. In contrast, DL is flexible with respect to the choice of track model, but it normally incorporates continuous polynomial models (e.g., constant velocity, constant acceleration, etc.), or piecewise polynomial models, for target trajectories. Another difference between PMHT and DL is the choice of optimization criterion. Whereas in PMHT the goal is maximization of the a posteriori probability [7,8,24,25] (MAP), in DL development in this paper the goal is maximum likelihood estimation. Finally, as we stated above, the DL tracker is naturally structured for concurrent classification and tracking when features are available.

The DL tracker is a Track-Before-Detect algorithm, it simultaneously performs data detection, association, and track estimation; it uses either unthresholded sensor data or thresholded data with a very low threshold. This increases the measurement data set by orders of magnitude, yet the DL tracker can handle the large amount of data, because of its linear (non-combinatorial) complexity. It operates on measurement data over several frames in a batch algorithm to obtain a track estimate. The Maximum Likelihood–Probabilistic Data Association (ML-PDA) is also a track-before detect tracker. ML-PDA is a parameter estimation technique which assumes deterministic target motion (no process noise). Originally developed in 1990 [32], it was later enhanced by incorporating measurement amplitude as a feature into the ML-PDA likelihood function [33]. Like the DL tracker ML-PDA is appropriate for very low observable targets, however, it has suffered from high computational complexity. It has been significantly improved recently [34,35], and later in the paper we compare its reported performance to the DL tracker.

When applied to a benchmark multi-target tracking problem, it was found that the computational cost of PMHT has roughly the same order of magnitude as the cost of MHT and JPDAF [7]. Due to similarity to PMHT, the DL tracker would most likely have a similar computational cost when applied to this problem. However, as mentioned above, the computational cost of the DL tracker scales only linearly with increasing numbers of sensors, whereas the cost of MHT scales exponentially. This cost advantage of the DL tracker is even more evident for the multi-sensor case [31,36].

The paper is organized as follows. In Section I the likelihood function is derived for joint tracking and association, Section II develop the DL equation for non-combinatorial maximization of the likelihood. Section III demonstrates performance examples and compares them with published results for JPDA, PMHT, and ML-PDA algorithms.

## I. Likelihood for Joint Tracking and Association

Consider k GMTI radar scans, resulting in n = 1… N measurements $\mathbf{X}(n) = (x_n, y_n, a_n, D_n)$, where $(x_n, y_n)$ are range and cross-range positions, $a_n$, is amplitude and $D_n$ is Doppler. Measurements n = 1… N include all the data used for estimation, what is often called batch. A likelihood of error measurement, $\mathbf{e}(n)$, is defined as follows. Error measurements are considered independent, therefore, according to the standard probabilistic procedure for independent data [37, 38], the joint probability density function (pdf), or likelihood is defined as

$$L(\{\mathbf{e}(n)\}) = \prod_{n \in N} \text{pdf}(\mathbf{e}(n)). \qquad (1)$$

Here $\Pi$ stands for a product over n = 1… N. A pdf($\mathbf{e}(n)$) is defined according to multiple hypotheses (note a difference in terminology, in MHT algorithm a hypothesis is every *association* between data and track; we call a hypothesis a track or clutter with unknown parameters). The measurement $\mathbf{X}(n)$ can originate from clutter, or from one of several moving objects, which are the alternative hypotheses. We number hypotheses h = 1…H; h=1 corresponds to clutter, h = 2…H correspond to H-1 tracks. According to the standard probabilistic procedure for alternative hypotheses [37,38],

$$\text{pdf}(\mathbf{e}(n)) = \sum_{h \in H} r(h)\, \text{pdf}(\mathbf{e}(n)|h), \qquad (2)$$

Here $\Sigma$ stands for a sum over h = 1… H; r(h) is an a priori probability that measurement n originates according to hypothesis h, and pdf($\mathbf{e}(n)$|h) is a conditional pdf for this hypothesis. Substituting eq.(2) into eq.(1), we obtain

$$L(\{\mathbf{e}(n)\}) = \prod_{n \in N} \sum_{h \in H} r(h)\, \text{pdf}(\mathbf{e}(n)|h). \qquad (3)$$

Note, a priori probabilities r(h) are unknown and have to be estimated from data same as other unknown parameters. This product of sums contains $H^N$ items, corresponding to all combinations of data and tracks or clutter (every data point could have originated according to any hypotheses). This huge number is the reason for combinatorial complexity of algorithms in the past. The above notation for conditional pdfs is a shorthand for pdf(**e**(n)| **M**$_h$(n)), where **M**$_h$(n) is a model predicting measurement **X**(n); if this measurement originates according to hypothesis h, then **M**$_h$(n) is an expected value of this measurement (when true parameter values are used),

**M**$_h$(n) = E{**X**(n)|h},                     (4)

and

**e**(n,h) = **X**(n) - **M**$_h$(n).                    (5)

We consider tracking short track segments, tracklets, along which velocities can be considered constant **V**$_h$ = ($V_{hx}$, $V_{hy}$). Correspondingly, the complete model is

**M**$_h$(**S**$_h$,n) = ($X0_h + V_{hx} t_n$, $Y0_h + V_{hy} t_n$, $a_h$, $D_h$).       (6)

Here parameters of the model, **S**$_h$ = ($X0_h$, $Y0_h$, $V_{hx}$, $V_{hy}$, $a_h$, $D_h$); ($X0_h$, $Y0_h$) model an original position, ($V_{hx}$, $V_{hy}$) model velocity, ($a_h$, $D_h$) model amplitude and Doppler; $t_n$, is the known time counted from the first scan. Also,

$V_{hx} = D_h$.                                  (7)

The unknown parameters include r(h), parameters of conditional pdf, such as standard deviations or covariances, and the total number of track-models. Conditional pdf for clutter we define as uniform,

pdf(**X**(n)|1) = 1/par_volume,             (8)

where par_volume is a volume of the parameter space, a product of ($S_{max} - S_{min}$) for all parameters. Conditional pdfs of tracks are defined as Gaussian; in view of hard boundaries $S_{max}$, $S_{min}$, this is an approximation; also parameters $a_h$ are not likely to follow Gaussian distributions. In our practical cases this approximate treatment has been sufficient,

$$\text{pdf}(\mathbf{e}(n,h)|h) = (2\pi)^{-2}(\det \mathbf{C}_h)^{-0.5}$$
$$\cdot \exp\{-0.5\, \mathbf{e}(n,h)^T \mathbf{C}_h^{-1}\, \mathbf{e}(n,h)\}. \quad (9)$$

We use diagonal covariance matrixes $\mathbf{C}_h$ = diag($\sigma_{x_h}^2$, □$\sigma_h^2$, □$\sigma_h^2$, □$\sigma_h^2$); and, $\sigma_{x_h}^2$ = □$\sigma_{D_h}^2$. This covariance matrix combines the uncertainties of the state and measurements.

## II. DYNAMIC LOGIC

Now we describe a procedure (an algorithm) to maximize the likelihood (1) while at the same time solving the association-assignment problem without combinatorial complexity. We call this procedure dynamic logic (DL) for reasons described in [39], [40]. The following algorithmic iterative equations (10) through (17) follow derivation in [14], where it is also related to the EM algorithm, and the local convergence proven.

DL is an iterative procedure, which starts with unknown values of model parameters and correspondingly large uncertainty of associations; this later requirement is achieved by setting standard deviation of parameters equal to one half of (max – min) value for this parameters (which are usually approximately known in an operation scenario). Taking these initial values of parameters, conditional probabilities eq.(9) are computed. Then association variables, f(h|n), are computed; they are defined similarly to posteriori Bayes probabilities (for shortness, we use indexes n, h instead of the corresponding data **X**(n) and models **M**$_h$(n).

$$f(h|n) = r(h)\, pdf(n|h) \,/\, \sum_{h' \in H} r(h')\, pdf(n|h'). \quad (10)$$

Although eq.(10) looks like posteriori Bayes probabilities, initially f(h|n) are not probabilities, since parameter values are incorrect; they can be called association variables or estimated probabilities of measurements n originating from tracks (objects or hypotheses) h.

The next step is to estimate parameters, using these estimated association probabilities. The equations for r(h),

$$r(h) = \sum_{n \in N} f(h|n)/N. \quad (11)$$

This gives an estimated average ratio of data points assigned to track h (or clutter h=1) to the total number of data points N. This and other parameter estimation equations look simpler with the following notation:

$$<...>_h = \sum_{n \in N} f(h|n)\,(...)_n. \quad (12)$$

Then eq.(11) can be rewritten as

$$r(h) = <1>_h / N. \quad (13)$$

Other parameter estimation equations, at each iteration, are computed as

$$a_h = <a_n>_h. \quad (14)$$

$$\begin{aligned}Y0_h <1>_h + V_{yh} <t_n>_h &= <Y_n>_h,\\ Y0_h <t_n>_h + V_{yh} <t_n^2>_h &= <Y_n\, t_n>_h.\end{aligned} \quad (15)$$

$$\begin{aligned}X0_h <1>_h + V_{xh} <t_n>_h &= <X_n>_h,\\ X0_h <t_n>_h + V_{xh} (<t_n^2>_h + c<1>) &= \\ &= <X_n\, t_n>_h + c <D_n>_h.\end{aligned} \quad (16)$$

Here, $c = \sigma_{x_h}^2 / \sigma_{\Phi_h}^2$. For the unknown parameters, $Y0_h$ and $V_{y_h}$, eq.(15) is a two-dimensional linear system of equations; similarly eq.(16) is a two-dimensional linear system of equations for $X0_h$ and $V_{x_h}$. Standard deviations for each parameter s are estimated, as follows:

$$\sigma_{hs}^2 = <(X_s(n) - M_{hs}(n))^2>_h \qquad (17)$$

DL consists in iterative computations of eqs.(10) through (17). While estimated parameters are far from true values, models do not match data, standard deviations are large, and associations f(h|n) are "flat": small numbers for many combinations of n and h (including incorrect ones), any data point has a nonzero assignment to any track (or clutter). Nevertheless, even with these poor initial associations, parameter values improve on every iteration according to a theorem proven in [14]: Likelihood (1) grows on every iteration and the DL procedure converges (local vs. global convergence is discussed below). As parameter values converge close to their true values, standard deviations converge to small values close to the sensor errors. Association variables converge close to true probabilities, close to 1 for n and h pairs corresponding to data n originating from object h, and to 0 otherwise (this last statement is true to the extent that the information contained in the data is sufficient for track separability and data association). This DL process from vague to crisp associations is characteristic of DL [14,[41]].

The computational complexity of the DL procedure described above is proportional to the number of data points and the number of tracks, const*N*H. The const here accounts for the number of iterations, and for complexity of procedures described by (10) through (17). Typical numbers are discussed in the next section. The principal theoretical moment is that this number is linear in $N$ and in $H$, rather than combinatorial, $\sim H^N$ like in MHT.

The number of tracks is estimated as follows. The algorithm starts with 1 *active* track model, which parameters are updated from iteration to iteration. In addition, the algorithm keeps one (or several) *dormant* track, which parameters are not updated, except for r(h). On the 1$^{st}$ iteration all standard deviations are large and the track and clutter models have low but nonzero associations to all data points. After few iterations the active track standard deviations become smaller, it is stronger associated with some returns and weaker with others. According to eq.(11), sum total associations of every data point n, $\Sigma_h f(h|n) = 1$; therefore some associations for dormant tracks grow. After r(h) for a dormant track exceeds a predetermined threshold, this track is activated and its parameters are updated. Similarly, if r(h) falls below the threshold, the track is eliminated. In this way as many tracks are activated as justified by the data. This procedure may lead to too many active tracks. When standard deviations approach sensor error values and association variables approach 0s and 1s, extra tracks tend to converge either on top of each other, or to one or two data points, these are pruned. Upon convergence (likelihood eq.(1) increase between iterations become less than a threshold), a detection measures is computed for each track; it is defined as a local log-likelihood ratio computed using returns within two standard deviations {n'} of each track:

$$LLR(h) = \sum_{n' \in N'} [\ln pdf(n'|h) - \ln pdf(n'|1)]. \qquad (18)$$

Tracks with LLR(h) exceeding a predetermined threshold are declared detections.

Convergence of this iterative procedure to a local maximum of similarity measure (1), as mentioned, was proven in [14]. Such local convergence usually occurs within relatively few iterations; a typical example in the next section took 20 iterations. Since similarity is a highly non-linear function, regular convergence to the global maximum can not be expected. The local rather than global convergence sometimes presents an irresolvable difficulty in many applications. In the presented method, this problem is resolved in several ways. First, the large initial standard deviation of the similarity measure smoothes local maxima. Second, tracks are pruned and activated as needed. Therefore if a particular real track is not "captured" after few iterations, it will be captured at a later iteration, after a track-model activation. Third, if a spurious track is declared detected, or a real track is missed, these errors will be self-corrected at a later stage of a system operation, when detected track segments or tracklets are connected into longer tracks (system operation procedures are beyond the current communication).

## III. Tracking example and ROC

An application example of the above described DL tracker is illustrated in Fig. 1, where detection and tracking are performed for targets below the clutter level. Fig. 1(a) shows true track positions in a 0.5km * 0.5km data set, while Fig. 1(b) shows the actual data available for detection and tracking. In this data, the target returns are buried in the clutter, with signal-to-clutter ratio of about –2dB for amplitude and –3dB for Doppler. Let us emphasize, target returns are lower than clutter returns and cannot be visible, this does not depend on renormalizing image brightness. Here, the data are displayed such that all six revisit scans are shown superimposed in the 0.5km * 0.5km area, 500 pre-detected signals per scan, and the brightness of each data sample is proportional to its measured Doppler value. Figs. 1(c)-1(h) illustrate the dynamics of the algorithm as it adapts during increasing iterations; the brightness is proportional to association variables, which for this display purpose are computed not just for **X**(n) but for all pixels (resulting in a smooth image shape). Only association variables for active track models are shown.

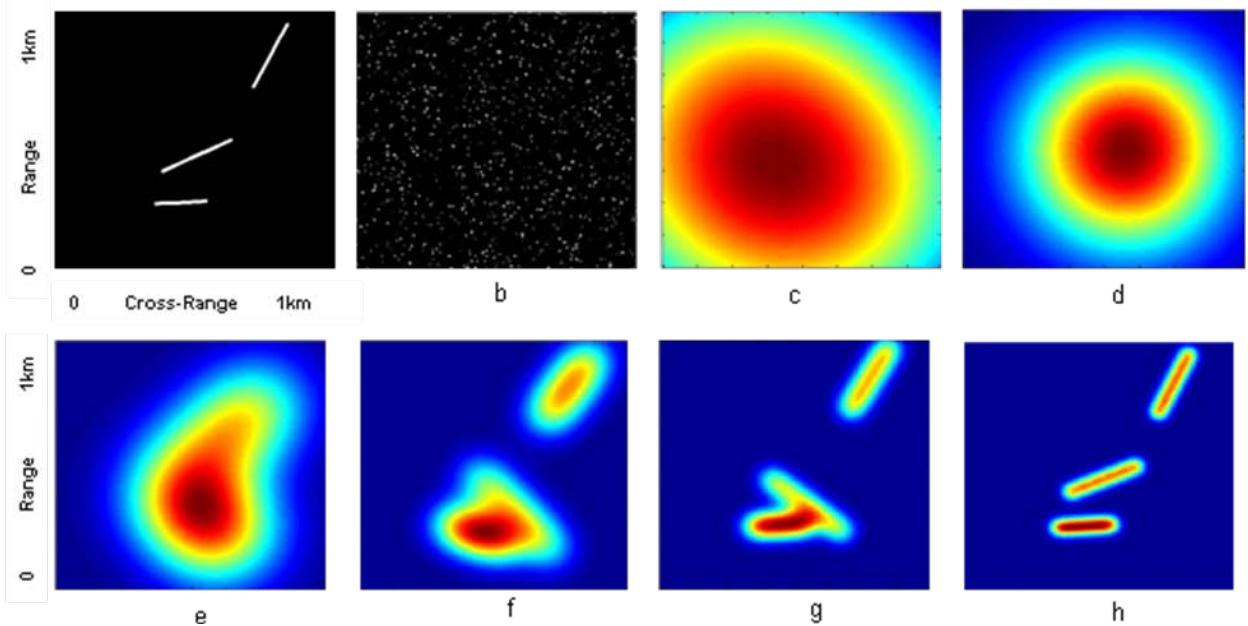

Fig. 1. Detection and tracking three targets in clutter using DL: (a) true track positions in 0.5km * 0.5km data set; (b) actual data available for detection and tracking. DL iterations are illustrated in (c) – (h), where (c) shows the initial, uncertain model and (h) shows the models upon convergence after 20 iterations. Note the close agreement between the converged models (h) and the truth (a).

Fig. 1(c) shows the initial vague track-model, and Fig. 1(h) shows track-models upon convergence at 20 iterations. Between (c) and (h) the DL tracker automatically decides how many track-models are needed to fit the data, and simultaneously updates the track parameters and association variables. There are two *types* of models: one uniform model describing clutter (it is not shown), and linear track-models, which uncertainty changes from large (c) to small (h). In (c) and (d), the DL tracker fits the data with one model, and uncertainty is somewhat reduced. Between (d) and (e) the DL tracker uses more than one track-model and decides that it needs two models to 'understand' the content of the data. Fitting with 2 tracks continues until (f); between (f) and (g) a third track is added. Iterations stop at (h), when similarity stops increasing. Detected tracks closely correspond to the truth (a).

In this example, target signals are below clutter. A single scan does not contain enough information for detection. Detection should be performed concurrently with tracking, using several radar scans, and six scans are used. In this case, a standard multiple hypothesis tracking, evaluating all tracking association hypothesis, would require about $10^{5000}$ operations, a number too large for computation. Therefore, existing tracking systems require strong signals, with about a 15 db signal-to-clutter ratio [1]. DL successfully detected and tracked all three targets and required only $10^6$ operations, achieving about 18 dB improvement in signal-to-clutter sensitivity.

A detailed characterization of performance requires operating curves (ROC), plots of probability of detection vs. probability of false alarm, computed for various signal-to-clutter ratios, densities of targets, target velocities, and other scenario parameters. Such detailed characterization is beyond the scope of this communication. Instead, Fig.2 illustrates three ROCs for selected parameter values.

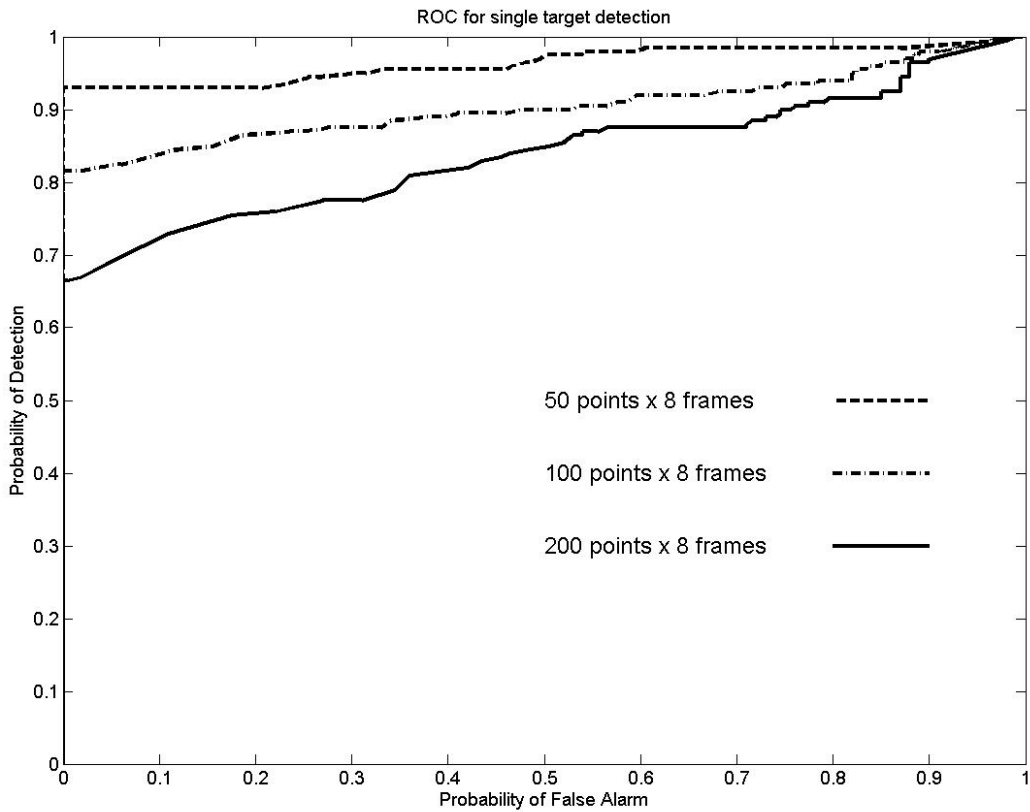

Fig. 2. Three ROC curves for different clutter levels: 50, 100, and 200 pre-detected signals per frame. 8 frames are used (total of 400, 800, and 1600 clutter signals per 8 target signals). Signal to clutter ratio, S/C, is defined as a signal strength divided by standard deviations taken as a sum of clutter and target standard deviations: $S/C = [(\sigma_C + \sigma_T)]$. S/C is 1.7 for amplitude and 2.0 for Doppler.

Results illustrated in Fig. 2 can be compared to PMHT and PDAF algorithm performance reported in [42] in Figs. 1 and 2. The reported performance for PDAF and PMHT is well beyond the DL tracker presented here: PDAF and PMHT performance with 10 clutter measurements per scan is below the performance reported here in Fig. 2 for 200 measurements per scan; PDAF and PMHT performance for 50 measurements per scan is significantly worse. Similar results for PMHT have been reported in [43]. These results are similar to those reported earlier in [44]. Preliminary performance of MLPDA algorithm (for track maintenance) have been reported in [34,35]; in these publications, Table III, Fig.4, the average number of false detections per surveillance region is 57 to 285; the surveillance region comprises 80 frames; the track batch size is 7 frames, which comparable to 8 frames in Fig.2 above. The total surveillance region (the area in x and y) is approximately two orders of magnitude larger than the batch size. In addition, PMHT and PDAF track initiation performance is significantly worse; the performance reported above in Fig. 2 for up to 200 clutter measurements per scan significantly exceeds other existing algorithms

## IV. Conclusion

The paper presented a maximum likelihood solution for tracking in clutter, while avoiding combinatorial complexity.

Future research will include feature-added tracking when - in addition to amplitude, position, and velocity - other characteristics of received signals are also used for improved associations between signals and track models. DL can naturally incorporate this additional information. Since association neural weights in DL are functions of object models (1) any object feature can be included into the models and will be used for signal-model associations.

Other sources of information can be included. For example, coordinates of roads can be easily incorporated into the DL procedure. For this purpose road positions should be characterized by a probability density, depending on the known coordinates and expected errors. Then similarities (2) can be modified by multiplying them by the probability densities of roads.

## Acknowledgement

The authors are thankful to AFOSR PM Dr. J. Sjogren for supporting part of this research. LP is thankful to participants of the ECE Colloquium, University of Connecticut at Storrs for detailed discussion.

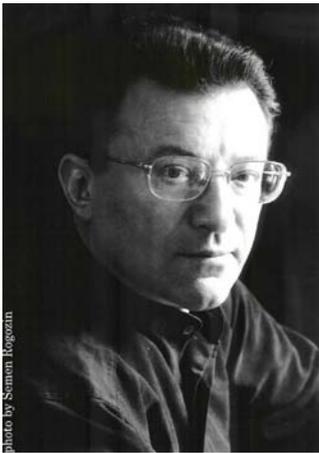**Leonid Perlovsky** (M'86-SM'81) Ph.D.'74 in theoretical and mathematical physics, Institute for Mathematics, Novosibirsk and Joint Institute for Nuclear Research, Dubna, Moscow, Russia.

He is Visiting Scholar at Harvard University, Cambridge, MA and Principal Research Physicist and Technical Advisor at the Air Force Research Laboratory, Hanscom AFB, MA, Program Manager for detection, tracking, sensor fusion, neural networks, modeling of the mind, language, cognition, culture, and other research projects. From 1985 to 1999, he served as Chief Scientist at Nichols Research. Previously Professor at Novosibirsk University and New York University, principal in commercial startups developing tools for text understanding, biotechnology, and financial predictions. His company predicted the market crash following 9/11 a week before the event, detecting activities of Al Qaeda traders. He delivered invited keynote plenary talks and tutorial lectures worldwide, published more then 380 papers, 11 book chapters, and authored 3 books, including "Neural Networks and Intellect," Oxford University Press, 2001 (currently in the $3^{rd}$ printing).

Dr. Perlovsky serves on the International Neural Network Society (INNS) Board of Governors, Chairs INNS Award Committee, Co-Chairs IEEE NN TC, Chairs TF on the Mind and Brain, Chairs IEEE Boston Computational Intelligence Chapter, organizes conferences on Computational Intelligence, serves on 5 Editorial Boards, including as Editor-in-Chief for "Physics of Life Reviews" (organized together with the Nobel Laureate I. Prigogine)  He received the IEEE Distinguished Member of Boston Section Award 2005; prestigious National and International awards including, The McLucas Award 2007, the top basic research award from the US Air Force and the INNS, and The Gabor Award, the top award from INNS for engineering achievements.

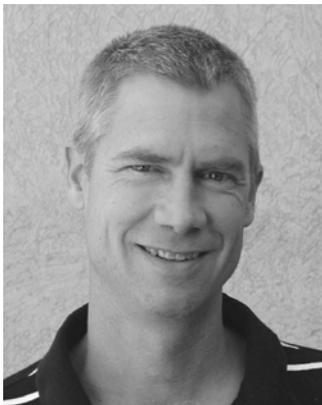**Ross Deming** (M'05) received the B.S. degree from Cornell University in 1985, the M.S. degree from the University of Vermont in 1993, and the Ph.D. degree from Northeastern University in 1996, all in electrical engineering.

He was previously employed at General Electric, Inc. and Teradyne, Inc., as an electrical engineer, and at Nichols Research Corp. and Witten Technologies, Inc., as a senior scientist.  Currently, he is a consultant for General Dynamics, Inc., where he is working on radar and optical signal processing problems for the Air Force Research Laboratory.

While at Witten Technologies (WTI), he was part of a small team that won, among other awards, the Wall Street Journal's 2004 Technology Innovation Award (1st place software category, honorable mention overall) for its technology that creates detailed images of objects and conditions underground using radar. He is also coauthor on a patent for this radar system.